Expert System Gradient Descent Style Training: Development of a Defensible Artificial Intelligence Technique

Jeremy Straub
Department of Computer Science
North Dakota State University

1320 Albrecht Blvd., Room 258
Fargo, ND 58108
p: +1 (701) 231-8196
f: +1 (701) 231-8255
e: jeremy.straub@ndsu.edu

**Abstract**

Artificial intelligence systems, which are designed with a capability to learn from the data presented to them, are used throughout society. These systems are used to screen loan applicants, make sentencing recommendations for criminal defendants, scan social media posts for disallowed content and more. Because these systems don't assign meaning to their complex learned correlation network, they can learn associations that don't equate to causality, resulting in non-optimal and indefensible decisions being made. In addition to making decisions that are sub-optimal, these systems may create legal liability for their designers and operators by learning correlations that violate anti-discrimination and other laws regarding what factors can be used in different types of decision making. This paper presents the use of a machine learning expert system, which is developed with meaning-assigned nodes (facts) and correlations (rules). Multiple potential implementations are considered and evaluated under different conditions, including different network error and augmentation levels and different training levels. The performance of these systems is compared to random and fully connected networks.

1. Introduction

Artificial intelligence systems have found use in numerous areas of society. These systems are used to screen loan applicants [1], to make sentencing recommendations for criminal defendants [2–4], to scan social media posts for disallowed content [5] and identify poster mental health issues [6], to make medical recommendations [7] and more. Many systems are designed with a capability to learn from the data presented to them as part of a supervised [8], semi-supervised [9] or unsupervised [10] training process. Through this learning process, the algorithm makes associations between input data and output results, thus identifying characteristics of inputs that are associated with characteristics of outputs.

However, because the systems don't assign meaning to the complex learned correlation network, they can learn associations that are based on correlations that are different from causality. They can also learn invalid relationships due to limited, skewed or erroneous training data. The creation of a bad model will, in most cases, result in non-optimal decisions being made. The potential for improperly informed or biased decisions, without the system or its developers even being aware that it is doing so, may be one reason why a survey by Araujo, et al. [11] found that the public is concerned about the use of artificial intelligence.

These systems may also learn correlations that violate anti-discrimination and other laws regarding what factors can be used in different types of decision making. O'Neil [12] notes that big data analytics has a "dark side", as autonomous decision systems can reinforce patterns of income discrimination, income equality and other societal inequities. D'Ignazio and Klein [13], similarly, note that autonomous data analysis systems may follow traditional power structures, based on the data that they learn from, thereby further disenfranchising the historically disenfranchised. Though perhaps well-meaning, machine learning systems can inadvertently become what Noble has aptly termed "algorithms of oppression" [14]. Spurious learning could also lead to civil liability [15,16].

This paper presents a potential solution to the problem of machine learning without the evaluation of associated correlation meaning. It considers the use of an expert system (see, [17] for more details about expert systems), which is developed with meaning-assigned nodes (facts) and correlations (rules) and the optimization of its decision making using machine learning techniques. Several different potential implementations are considered and evaluated under multiple operating conditions, including different network designs, error and augmentation (as compared to a perfect system) levels and different training levels. The performance of these approaches is compared. The proposed systems are also compared to systems using random and fully connected networks.

2. Background

This section considers prior work in key areas that provide a foundation for the current work presented herein. First, work on expert systems is reviewed. Then, prior work on gradient descent and other related learning techniques is presented.

2.1. Expert Systems

Some contend that expert systems were introduced in 1965 by Feigenbaum and Lederberg [18], who developed a system called Dendral, which separated knowledge storage from its processing engine and used a rule-based engine for problem solving [19].  The Dendral system was designed to formulate hypotheses for chemistry problems.  Others, though, consider the Mycin system, developed in the 1970s for medical applications, to be the first true expert system [19].

Classical expert systems used a rule-fact network for inference [17].  In the most basic form, facts can have the binary values of true or false and rules identify additional facts that can be asserted as true, based on two precursor facts being asserted as true.  The rules processing engine scans the knowledge base for rules whose preconditions are satisfied and runs them, setting the value of designated postcondition facts.  Expert systems have found use in numerous fields including medicine [20][21], power systems [22], facial feature identification [23], agriculture [24], education [25] and geographic information systems [26].

A variety of enhancements have been proposed to expert systems, including different optimization techniques [27] such as using genetic algorithms and principals related to neural networks.  Hybrid expert systems [28] that use optimization and other techniques, such as back-propagation, from neural networks and systems that are actual neural networks have been demonstrated.

Expert systems based on fuzzy logic have also been proposed.  These systems use fuzzy set concepts [29] and can represent the uncertainty present in fact values and rules.  Mitra and Pal [30] define a taxonomy for these fuzzy logic-based systems.  "Fuzzy expert systems" just use fuzzy sets as part of a normal expert systems, while "fuzzy neural network[s]" use fuzzy sets and a neural network [30].  When neural networks and expert systems are combined, with a neural network's connection weights being used to generate rules (instead of a human knowledge engineer), this is a "connectionist expert system" [30].  "Neuro-fuzzy expert systems" build on connectionist concepts, storing the knowledge base in a fuzzy neural network.  Several "neuro-fuzzy expert systems" have been developed including systems for diabetes therapy [31], autonomous vehicle obstacle avoidance [32], heart disease diagnosis [33], hypertension diagnosis [34], and software architecture evaluation [35].

Mitra and Pal [30], though, go beyond this basic concept and define a "knowledge-based connectionist expert system" which starts with "crude rules" that are stored as connection weights within a neural network and then training is used to produce refined rules.  A Google Scholar search did not find any references to implementations of systems of this type.  While this type of system could have a predisposition towards valid associations, based upon the initial "crude rules", training within the neural network could still cause the system to learn invalid and potentially problematic relationships.

2.2. Gradient Descent and Machine Learning

Gradient descent and, specifically, back-propagation techniques are used frequently with neural network implementations.  Gradient descent seeks to optimize inputs to find a minimum output, such as the error level value of training data and its result.  Backpropagation [36] changes weights throughout a neural network, iteratively, based on differences between the network-produced and target output values.

A variety of enhancements to these techniques have been proposed. These include techniques focused on speed enhancement [37], including those which introduce noise [38], use federated learning and momentum [39] and use evolutionary algorithms [40]. Other prior work has focused on system bias factors [41,42], initial condition sensitivity [43] and attack resilience [44,45]. Yet others have tried incorporating other artificial intelligence techniques for training, including genetic algorithms [46], Levenberg-Marquardt training algorithms [47], particle swarm optimization [48] and simulated annealing [48].

Techniques which use speculative approaches [49], spiking neural network concepts [50,51] and memory use optimization [52] have also been proposed. Kim and Ko [53] and (separately) Ma, Lewis and Kleijn [54] have proposed techniques which train neural networks while specifically avoiding backpropagation altogether.

Problematically, beyond the general concepts of gradient descent and backpropagation, these techniques largely depend on or are optimized for the particular structure used by neural networks. This makes them informative, but not directly useful, for other areas of potential application.

3. System Description

This paper proposes to conceptually merge an expert system and neural network. Unlike the methods of using neural networks to generate rules for expert systems, discussed by Mitra and Pal [30], it proposes an approach to applying the concepts of gradient descent and backpropagation on the expert system rule network itself.

To this end, the system proposed herein is comprised of a typical expert system engine that processes rules in a forward fashion. A training module, that operates largely independent of the expert system engine (the engine is used as part of training to determine the output of the rule-fact network) is used to optimize the rule weightings. This module is described in Section 4.

An expert system that supports the concept of partial membership and ambiguity, where facts can have a probabilistic or partial membership value between 0 and 1 instead of simply being true or false, is used for this work. In line with this, rules are more complex than simply requiring that two facts be true, as a precondition, and asserting a third, if the precondition is met. Instead, rules have weighting values for the comparative impact of both input facts on the value of the output fact. These weightings must be between 0 and 1 and their sum must be 1.

For the purposes of experimentation, the expert system is tasked with determining the value of a target final fact. During each system run, a rule-fact network is created randomly. The number of rules and facts are user-defined simulation parameters. When rules are created, facts are selected at random as input and output facts. Rules are checked against the existing ruleset to prevent duplication. During a run, an initial fact and final fact are randomly selected from all of the facts available in the network. The initial fact has an initial value of 0.99 assigned and the final fact's value is recorded at the end of the run. A run ends when no facts are modified during a particular iteration of operations. Runs end immediately if the final fact starts with a value satisfying run completion.

To test the effect of different training approaches, the performance of the system under each experimental condition was evaluated 1,000 times. Evaluation runs that do not complete (due to connectivity and other issues with the randomly generated networks) and runs that complete immediately are excluded from error-result and processing time averages. All runs were performed on the same workstation with a tenth generation Intel i7 processor and 16 GB of RAM.

4. Gradient Descent Inspired Training Mechanism

The proposed system requires a new algorithm for training, as its configuration is significantly different than the configuration of a typical neural network that backpropagation is designed to work with. Figure 1 (left) shows a typical neural network and its interconnections. Notably each node in a given layer is connected to all nodes in the previous and subsequent layers; however, nodes within a given layer are not connected. Additionally, nothing is connected to a node in a layer that is not directly adjacent to the current layer.

Because of this configuration, every other node in the network – besides the other nodes in the output layer – potentially contributes to the value of each output node. This is depicted in Figure 1 (right). This is dramatically different from the configuration of the proposed system where no concept of layers exists and any node can, potentially, be connected to any other node. However, in many cases, these connections are sparse and only a limited number of nodes contribute to the value of a particular target node.

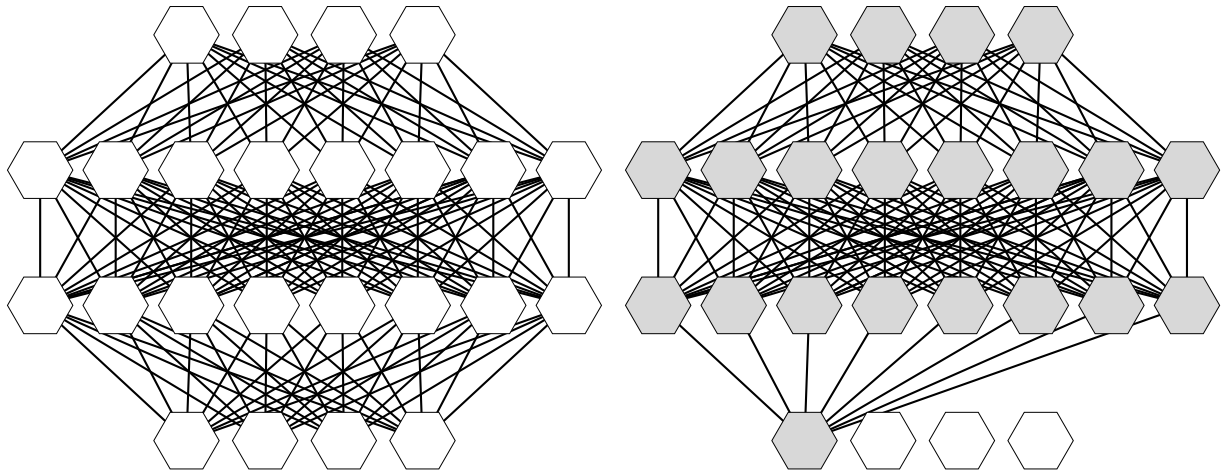

Figure 1. Fully connected 4-layer neural network with two hidden layers (left) and network showing what nodes and segments contribute to the value of a given output node (right).

The comparative number of facts and rules in the network as well as (for the randomly created networks used for testing) random chance determines how many and what facts and rules can contribute to a given target fact. This is visually depicted in Figure 2. Notably, while Figure 2 shows the facts and rules arranged in a layer-like manner for readability, no such logical or physical organization exists within the system.

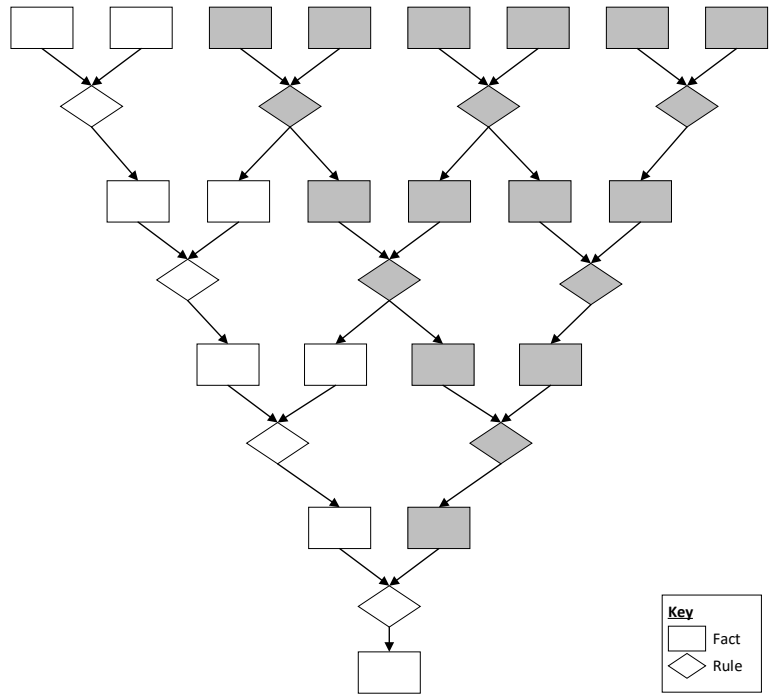

Figure 2. Showing rules and facts contributing to the value of a given fact in an expert system.

Given this, the linear algebra used for neural network backpropagation is not suitable for the learning expert system. Instead, a new model for distributing change is proposed and evaluated herein. This model identifies all rules that directly or indirectly contribute to the target node and distributes a portion of the difference between the truth node (input data used for training) value and the network-under-training's computed node value to the weightings of these rules. The percentage of the difference that is split between all contributing nodes is determined by a user-configurable velocity value.

In addition to the user-configurable velocity value setting, the number of training epochs, the network configuration type, the number of facts and rules, and the training model that are used are all user configurable. These settings form the bases for the experiments and experimental conditions described in Sections 5 to 7 and are described, more fully, at the point that they are first discussed.

For each training epoch, the network is run in forward-mode and the contributing rules to the designated target fact are identified. The contribution of each rule is determined by multiplying rules that must pass through other rules for their impact by the percentage impact of the relevant input of the intervening rules. Once the level of error between the truth and algorithm-generated network is determined, a percentage of this error (determined by the velocity setting value) is applied to each contributing rule's weightings based on its level of contribution and the value of its facts. Depending on the change that is required to the target node value, additional or reduced weight is, appropriately, given to the higher and lower values' input facts weightings. Subsequent epochs repeat this process.

5. Network Implementation Evaluation

In this section, the defensible artificial intelligence technique is explored via the evaluation of the benefits of its key characteristics. First, the impact of the network design is assessed. Then, the impact of training on the network is assessed.

*5.1. Impact of Network Types*

The assessment of the proposed technique starts with an evaluation of the impact of the network itself. For the purposes of this experiment, networks were randomly generated, as previously described, with multiple characteristics. In all cases, a base network was generated that served as the truth network. For the base condition (as shown in Tables 1 to 3), this network was replicated with its rule weightings reset and no other changes made. The augmented conditions added additional rules to the network corresponding to the percent (1%, 5%, 10%, 25% and 50%) shown and reset the rule weightings. The error networks changed a percent of the rules corresponding to the value shown (10%, 25% and 50%) in the table and reset the rule weightings. The fully connected (FC) network was generated by creating interconnecting rules between all facts. The random network was created by regenerating the network's rules without regard to the rules in the base network.

All of the experimental conditions were trained and evaluated 1,000 times and the averages of the runs are shown in the tables. Note that only runs that successfully complete are included in the processing time and result averages. Runs where the final fact cannot be asserted or starts asserted are not included. A velocity of 0.1, path-based training – where a single path is used for all training activities, and a consistent set of facts (for each individual run) were used. The data in Tables 1 to 3 was generated with a network 11 facts. In all cases, except for fully connected, 11 rules were also created. The FC network created rules interconnecting all facts. The limited network size used is due to the requirements of the FC network and the number of rules required to fully connect the nodes. An 11-node network requires hundreds of rules and, thus, requires a significant amount of memory and processing time. The overall training and testing time of the 1,000 tests for the FC network is 46,570 milliseconds as opposed to 17,009 milliseconds for the base network. This time includes all run setup activities. Individual run times for the FC network take orders of magnitude longer than for other network types. The base network average runtime, for example, was 299 tics versus 169,338 tics for the fully connected network.

Table 1 presents the average results of the truth network and the algorithm-generated network, after one training epoch. Table 2 compares the results of the truth network and the algorithm-generated network in terms of the mean and median error values from the 1,000 runs. Finally, Table 3 illustrates the performance of the various network types in terms of higher-error level and lower-error level networks. As backpropagation / gradient descent [55] and neural networks [56] are known to suffer from issues related to local minima, the different impact of approaches on higher and lower error-level runs is specifically considered throughout this paper.

Table 1. Comparison of average results for an 11-node network and 1000 runs with 1 epoch of training.

|  | Truth Network | Algorithm Network |
| --- | --- | --- |
| Base | 0.577 | 0.561 |
| Fully Connected | 0.521 | 0.565 |
| Random | 0.507 | 0.579 |
| Augmented 1% | 0.576 | 0.569 |
| Augmented 5% | 0.576 | 0.578 |
| Augmented 10% | 0.579 | 0.581 |
| Augmented 25% | 0.582 | 0.587 |
| Augmented 50% | 0.566 | 0.567 |
| Error 10% | 0.563 | 0.571 |
| Error 25% | 0.594 | 0.603 |
| Error 50% | 0.582 | 0.583 |

Table 2. Comparison of average error levels for an 11-node network and 1000 runs with 1 epoch of training.

|  | Avg Abs Error | Median Abs Err |
| --- | --- | --- |
| Base | 0.094 | 0.068 |
| Fully Connected | 0.315 | 0.280 |
| Random | 0.275 | 0.256 |
| Augmented 1% | 0.094 | 0.065 |
| Augmented 5% | 0.099 | 0.067 |
| Augmented 10% | 0.090 | 0.060 |
| Augmented 25% | 0.091 | 0.067 |
| Augmented 50% | 0.098 | 0.069 |
| Error 10% | 0.091 | 0.064 |
| Error 25% | 0.099 | 0.070 |
| Error 50% | 0.082 | 0.059 |

Table 3. Comparison of average higher and lower error level runs for an 11-node network and 1000 runs with 1 epoch of training.

|  | Avg of > .1 Error | Avg of < .1 Err | % < .1 Err |
| --- | --- | --- | --- |
| Base | 0.183 | 0.042 | 0.629 |
| Fully Connected | 0.411 | 0.030 | 0.251 |
| Random | 0.336 | 0.059 | 0.220 |
| Augmented 1% | 0.194 | 0.038 | 0.643 |
| Augmented 5% | 0.203 | 0.040 | 0.640 |
| Augmented 10% | 0.193 | 0.040 | 0.672 |
| Augmented 25% | 0.183 | 0.040 | 0.641 |
| Augmented 50% | 0.189 | 0.043 | 0.624 |
| Error 10% | 0.182 | 0.040 | 0.637 |
| Error 25% | 0.201 | 0.039 | 0.628 |
| Error 50% | 0.178 | 0.040 | 0.697 |

The impact of the rule-fact network itself is readily apparent from the data presented in Table 2. The base network (where the rule-fact structure is the same, but network weightings have been resent with one training epoch of correction) performs demonstrably better than the FC and random networks. The augmented and error networks also perform better than the random and FC networks, which exhibit approximately triple the error of other approaches. Notably, the FC and random networks, as shown in Table 3, have a much higher level of high-error runs (approximately 75% of runs, in both cases) than other network types (less than 40% in other cases). However, the FC and random networks have a similar average error level for the lower-error level runs as most other network types, with the FC network actually performing best, in terms of this metric.

From the foregoing, the impact of the network structure, alone, is very clear and, thus, is a notable contribution of the proposed technique, in addition to the additional accuracy provided through the rules' fact contribution weightings. Notably, the FC network was able to be readily trained to perform well, but at a much higher processing cost, and with a greater risk of becoming trapped in a local minima.

While an 11-node network was used to facilitate the comparison of networks' performance to the FC network (which grows in complexity by at least an order of magnitude for every additional node beyond 10), 100-rule/100-fact networks will be used for most other analysis in this paper (the impact of different network rule and fact combinations is also considered, later). For the purposes of comparison, Tables 4 to 6 present data similar to that presented in Tables 1 to 3 for 100-rule/100-fact networks. As is to be expected, the processing time of the 100-node networks is greater than the 11-node networks. For the entire 1,000 runs of training and testing, the 100-node network took 21,635 milliseconds versus 17,009 milliseconds for the 11-node network. Notably, these times include setup, processing and evaluation activities, in addition to just the training times, making the differences between the values (as opposed to the values themselves) of interest. It is also worth noting that the 1-epoch-of-training networks process faster than networks with more training, which become more complex as weightings are optimized.

Table 4. Comparison of average results for a 100-node network and 1000 runs with 1 epoch of training.

|               | Truth Network | Algorithm Network |
|---------------|---------------|-------------------|
| Base          | 0.514         | 0.513             |
| Random        | 0.403         | 0.522             |
| Augmented 1%  | 0.532         | 0.537             |
| Augmented 5%  | 0.512         | 0.517             |
| Augmented 10% | 0.518         | 0.518             |
| Augmented 25% | 0.489         | 0.480             |
| Augmented 50% | 0.512         | 0.514             |
| Error 10%     | 0.497         | 0.506             |
| Error 25%     | 0.521         | 0.513             |
| Error 50%     | 0.508         | 0.519             |

Table 5. Comparison of average error levels for a 100-node network and 1000 runs with 1 epoch of training.

|  | Avg Abs Error | Median Abs Err |
|---|---|---|
| Base | 0.090 | 0.064 |
| Random | 0.251 | 0.224 |
| Augmented 1% | 0.084 | 0.057 |
| Augmented 5% | 0.085 | 0.063 |
| Augmented 10% | 0.092 | 0.077 |
| Augmented 25% | 0.082 | 0.062 |
| Augmented 50% | 0.082 | 0.058 |
| Error 10% | 0.085 | 0.065 |
| Error 25% | 0.077 | 0.057 |
| Error 50% | 0.091 | 0.059 |

Table 6. Comparison of average higher and lower error level runs for a 100-node network and 1000 runs with 1 epoch of training.

|  | Avg of > .1 Error | Avg of < .1 Err | % < .1 Err |
|---|---|---|---|
| Base | 0.180 | 0.041 | 0.647 |
| Random | 0.313 | 0.058 | 0.242 |
| Augmented 1% | 0.179 | 0.040 | 0.685 |
| Augmented 5% | 0.175 | 0.042 | 0.674 |
| Augmented 10% | 0.177 | 0.045 | 0.647 |
| Augmented 25% | 0.180 | 0.041 | 0.702 |
| Augmented 50% | 0.175 | 0.038 | 0.676 |
| Error 10% | 0.173 | 0.041 | 0.667 |
| Error 25% | 0.174 | 0.039 | 0.713 |
| Error 50% | 0.194 | 0.038 | 0.659 |

With this level of rules and facts, the random network still performs notably worse than the other network types; however, the other networks (with errors or augmentations), perform roughly equivalently. With only one training Epoch, error for the non-random networks ranges from approximately 8% to 9% (as compared to the approximately 25% error for the random network with one epoch of training). These results show that the proposed approach exhibits robustness with regards to having additional and erroneous nodes. This is important, as it means that small errors in network creation (either by humans or by a future automated technique) may not, in many cases, have a large impact on the efficacy and accuracy of the system.

*5.2. Impact of Training Epochs*

Next, the impact of different levels of training is assessed. The results from training levels ranging from 1 epoch to 1000 epochs are presented in Tables 7 to 9. An epoch, for this system, is defined as a single run of the expert system and the distribution of a fraction of the difference (specified by the velocity setting) between the network-under-training and truth network output values. As expected, the level of time required for training increases significantly between training levels. As an example, the overall run time for 1,000 runs of training, processing and evaluation increases from 159,991 milliseconds for 10 epochs of training to 429,710 milliseconds for 25 epochs of training to 1,843,462 milliseconds for 100 epochs of training.

Table 7. Comparison of average results for a 100-node network and 1000 runs with different levels of training.

|  | Truth Network | Algorithm Network |
|---|---|---|
| 1 Epoch | 0.514 | 0.513 |
| 10 Epochs | 0.506 | 0.510 |
| 25 Epochs | 0.509 | 0.512 |
| 50 Epochs | 0.508 | 0.513 |
| 100 Epochs - Base | 0.522 | 0.521 |
| 250 Epochs | 0.520 | 0.536 |
| 500 Epochs | 0.506 | 0.528 |
| 1000 Epochs | 0.527 | 0.548 |

Table 8. Comparison of average error levels for a 100-node network and 1000 runs with different levels of training.

|  | Avg Abs Error | Median Abs Err |
|---|---|---|
| 1 Epoch | 0.090 | 0.064 |
| 10 Epochs | 0.094 | 0.065 |
| 25 Epochs | 0.083 | 0.061 |
| 50 Epochs | 0.086 | 0.056 |
| 100 Epochs - Base | 0.083 | 0.050 |
| 250 Epochs | 0.091 | 0.048 |
| 500 Epochs | 0.088 | 0.041 |
| 1000 Epochs | 0.091 | 0.038 |

Table 9. Comparison of average higher and lower error level runs for a 100-node network and 1000 runs with different levels of training.

|  | Avg of > .1 Error | Avg of < .1 Err | % < .1 Err |
|---|---|---|---|
| 1 Epoch | 0.180 | 0.041 | 0.647 |
| 10 Epochs | 0.194 | 0.041 | 0.656 |
| 25 Epochs | 0.173 | 0.041 | 0.688 |
| 50 Epochs | 0.183 | 0.038 | 0.668 |
| 100 Epochs - Base | 0.192 | 0.036 | 0.702 |
| 250 Epochs | 0.216 | 0.031 | 0.678 |
| 500 Epochs | 0.222 | 0.027 | 0.684 |
| 1000 Epochs | 0.227 | 0.023 | 0.665 |

While the average error fluctuates with different levels of training and does not show a clear pattern, the median of the absolute value, shown in Table 8, shows a clear decrease in error as more training is performed. This is explained when considering the average of the higher-error and lower-error runs, shown in Table 9. The lower-error runs show a pattern of decreasing error level, while a trend of increased average error is shown in the higher-error runs. Thus, it appears that increased training tends to increase the error of networks that may be trapped in a local minima, while further reducing the error levels of networks not experiencing this issue. This would appear to be consistent with prior work by Bianchini, Gori and Maggini [56], who noted a similar phenomenon when studying neural networks.

6. Impact of Network Implementation on Training Results

Having evaluated the comparative impact of different networks with only one epoch of training and the impact of different levels of training, focus now turns to comparing the different potential network configurations with more training (100 epochs) and comparing these networks to their performance with one epoch of training.

*6.1. Comparison of Base Versus Random Network with Additional Training*

Tables 10 to 12 present a comparison of the base (same structure as the truth network with reset rule input fact weightings) and random networks with 100 epochs of training. Notably the random network had over double the error level, both in terms of average and median values, and less than half the number of lower-error level runs. The 1,000 runs of the base and random networks did not have a notable difference in runtime, as would be expected as the random network is similarly complex to the base network.

Table 10. Comparison of average results for a base and random network with 100 epochs of training.

|  | Truth Network | Algorithm Network |
|---|---|---|
| Base | 0.522 | 0.521 |
| Random | 0.439 | 0.522 |

Table 11. Comparison of average error levels for a base and random network with 100 epochs of training.

|  | Avg Abs Error | Median Abs Err |
|---|---|---|
| Base | 0.083 | 0.050 |
| Random | 0.226 | 0.201 |

Table 12. Comparison of average higher and lower error level runs for a base and random network with 100 epochs of training.

|  | Avg of > .1 Error | Avg of < .1 Err | % < .1 Err |
|---|---|---|---|
| Base | 0.192 | 0.036 | 0.702 |
| Random | 0.302 | 0.045 | 0.294 |

The additional training of the random network was not highly productive, producing an approximate 10% decrease in both average and median error levels (comparing the data in Table 11 to Table 5). This is similar to the impact of additional training on the base network.

Demonstrably, additional training does not improve the performance of the random network to be equivalent to the base network, showing the impact of the accuracy of the rule-fact network.

*6.2. Comparison of the Performance of Different Sizes of Fully Connected and Base Networks with Greater Training*

Fully connected networks could, inherently, be trained to have the pattern of any other network configuration, simply by setting inapplicable rule fact-contribution weights to 0. Thus, if they could be efficiently trained and run, they would be an ideal solution as they could learn the pattern of the truth network using a gradient descent or other similar process and would not require manual network creation or the development of another algorithm to create the network to train.

Thus, the performance of fully connected (FC) and non-fully connected (NFC) base networks is compared in this section. These results are presented in Tables 13 to 15.

Table 13. Comparison of average results for different sizes of fully connected and base networks with 100 epochs of training.

|  | Facts / Rules | Truth Network | Algorithm Network |
|---|---|---|---|
| FC | 5 | 0.602 | 0.630 |
| FC | 7 | 0.555 | 0.610 |
| FC | 9 | 0.524 | 0.572 |
| FC | 10 | 0.548 | 0.582 |
| FC | 11 | 0.524 | 0.574 |
| NFC | 5 | 0.652 | 0.640 |
| NFC | 7 | 0.617 | 0.598 |
| NFC | 9 | 0.615 | 0.613 |
| NFC | 10 | 0.561 | 0.571 |
| NFC | 11 | 0.578 | 0.584 |

Table 14. Comparison of average error levels for different sizes of fully connected and base networks with 100 epochs of training.

|  | Facts / Rules | Avg Abs Error | Median Abs Err |
|---|---|---|---|
| FC | 5 | 0.305 | 0.220 |
| FC | 7 | 0.319 | 0.250 |
| FC | 9 | 0.321 | 0.250 |
| FC | 10 | 0.319 | 0.271 |
| FC | 11 | 0.312 | 0.257 |
| NFC | 5 | 0.141 | 0.052 |
| NFC | 7 | 0.124 | 0.062 |
| NFC | 9 | 0.103 | 0.044 |
| NFC | 10 | 0.113 | 0.049 |
| NFC | 11 | 0.114 | 0.054 |

Table 15. Comparison of average higher and lower error level runs for different sizes of fully connected and base networks with 100 epochs of training.

|      | Facts / Rules | Avg of > .1 Error | Avg of < .1 Err | % < .1 Err |
|------|---------------|-------------------|-----------------|------------|
| FC   | 5             | 0.461             | 0.030           | 0.362      |
| FC   | 7             | 0.443             | 0.030           | 0.299      |
| FC   | 9             | 0.439             | 0.036           | 0.293      |
| FC   | 10            | 0.416             | 0.033           | 0.254      |
| FC   | 11            | 0.409             | 0.041           | 0.263      |
| NFC  | 5             | 0.346             | 0.029           | 0.648      |
| NFC  | 7             | 0.289             | 0.033           | 0.645      |
| NFC  | 9             | 0.282             | 0.034           | 0.721      |
| NFC  | 10            | 0.269             | 0.030           | 0.653      |
| NFC  | 11            | 0.282             | 0.036           | 0.682      |

It is notable that the training time for the FC networks increases significantly with each additional fact that is included. The 1,000 runs of the 10-node FC network, for example, requires 775,977 milliseconds as opposed to the nearly double 1,377,366 milliseconds required for the 11-node FC network. For purposes of comparison, the NFC processing time increases from 18,884 to 19,961 milliseconds between 10 and 11 nodes. The 11-fact network, where this experiment stops at, requires hundreds of rules to fully interconnect the fact network. Training, processing and evaluating the 11-fact FC networks takes almost as much time as for the 100-fact base (NFC) networks. Additionally, with similar training, the FC networks have an average error that is double or more the NFC networks and a median error level that is five to ten times as high.

What is perhaps most notable, about training the FC networks, is that the lower-error level runs reach approximately the same average error as the NFC lower-error level runs; however, only about half as many runs fall into this category. The error of the higher-error level runs is large, on the other hand, averaging just under half of the possible value range of the target fact. This shows that, as is expected, the FC networks can be trained; however, they are far more computationally expensive to train and operate and far more likely to end up in a local minima situation or an invalid configuration, as compared to the NFC networks. While the operating time could potentially be reduced by pruning non-contributing and minimally contributing nodes, this approach wouldn't necessarily fix the error present in the higher-error level runs.

*6.3. Comparison of the Performance of Augmented and Error Networks with Additional Training and Additional Training on Network Types*

In Section 5.1, the performance of various network types with minimal (1 epoch) of training was assessed. Now, the different network types are compared with 100 epochs of training and the impact of the additional training is compared for each network type.

Tables 16 to 18 present the impact of additional training on the augmented network types. Tables 19 to 21 present the impact of additional training on networks with induced error.

Table 16. Comparison of average results for networks with different levels of augmentation with 100 epochs of training.

|                | Truth Network | Algorithm Network |
|----------------|---------------|-------------------|
| Base           | 0.522         | 0.521             |
| 1% Augmented   | 0.496         | 0.500             |
| 5% Augmented   | 0.479         | 0.499             |
| 10% Augmented  | 0.499         | 0.517             |
| 25% Augmented  | 0.517         | 0.537             |
| 50% Augmented  | 0.510         | 0.529             |

Table 17. Comparison of average error levels for networks with different levels of augmentation with 100 epochs of training.

|                | Avg Abs Error | Median Abs Err |
|----------------|---------------|----------------|
| Base           | 0.083         | 0.050          |
| 1% Augmented   | 0.081         | 0.051          |
| 5% Augmented   | 0.085         | 0.053          |
| 10% Augmented  | 0.104         | 0.067          |
| 25% Augmented  | 0.089         | 0.052          |
| 50% Augmented  | 0.090         | 0.061          |

Table 18. Comparison of average higher and lower error level runs for networks with different levels of augmentation with 100 epochs of training.

|                | Avg of > .1 Error | Avg of < .1 Err | % < .1 Err |
|----------------|-------------------|-----------------|------------|
| Base           | 0.192             | 0.036           | 0.702      |
| 1% Augmented   | 0.190             | 0.037           | 0.711      |
| 5% Augmented   | 0.197             | 0.038           | 0.705      |
| 10% Augmented  | 0.215             | 0.037           | 0.625      |
| 25% Augmented  | 0.208             | 0.036           | 0.692      |
| 50% Augmented  | 0.190             | 0.039           | 0.660      |

Table 19. Comparison of average results for networks with different levels of error with 100 epochs of training.

|           | Truth Network | Algorithm Network |
|-----------|---------------|-------------------|
| Base      | 0.522         | 0.521             |
| 10% Error | 0.502         | 0.499             |
| 25% Error | 0.506         | 0.526             |
| 50% Error | 0.497         | 0.513             |

Table 20. Comparison of average error levels for networks with different levels of error with 100 epochs of training.

|  | Avg Abs Error | Median Abs Err |
|--|---------------|----------------|
| Base | 0.083 | 0.050 |
| 10% Error | 0.080 | 0.051 |
| 25% Error | 0.087 | 0.048 |
| 50% Error | 0.084 | 0.049 |

Table 21. Comparison of average higher and lower error level runs for networks with different levels of error with 100 epochs of training.

| | Avg of > .1 Error | Avg of < .1 Err | % < .1 Err |
|---|---|---|---|
| Base | 0.192 | 0.036 | 0.702 |
| 10% Error | 0.190 | 0.036 | 0.715 |
| 25% Error | 0.209 | 0.037 | 0.710 |
| 50% Error | 0.189 | 0.035 | 0.679 |

As Tables 17 and 20 show, there is limited impact from augmenting the network or inducing error. The error levels fluctuate between augmentation and error levels, but do not show a clear pattern. Training, processing and evaluation time, similarly, fluctuates between the different sets of runs, but does not show a clear pattern. While the overall impact of augmentation and error is limited for a given trained path, this is likely due to compensation being caused within the network by the training process, which may result in the rule weightings and fact values of intervening rules and facts differing significantly from the truth network.

Comparing the 1-epoch of training and 100-epoch of training results, the augmented networks show a limited decrease in error (while showing some fluctuation). Similar limited benefit is seen with two of the error-induced networks, when comparing Tables 5 and 20. When comparing Tables 6 and 21, there is no clear pattern between the higher-error runs; however, the average error of the lower-error runs decreases for all three error levels.

Thus, while additional training provides some benefit for the base network configuration, it doesn't appear to provide benefit consistently for the augmented networks, but does reduce the median error of the error-induced ones. Of course, the additional epochs of training increases the time required for the training process.

*6.4. Comparison of the Impact of Different Velocity Levels on System Performance*

The velocity value determines how quickly change, based on the identified discrepancy between the network-under-training and the truth network, is applied to the network-under-training. This section assesses whether different velocity values impact the performance of the system. Tables 22 to 24 present data for six different velocity values, ranging from 0.01 to 0.50 (the base value used throughout other experiments is 0.10).

Table 22. Comparison of average results for different velocity levels.

| Velocity | Truth Network | Algorithm Network |
|---|---|---|
| 0.01 | 0.499 | 0.494 |
| 0.05 | 0.513 | 0.526 |
| 0.10 - Base | 0.522 | 0.521 |
| 0.15 | 0.505 | 0.510 |
| 0.25 | 0.502 | 0.527 |
| 0.50 | 0.516 | 0.548 |

Table 23. Comparison of average error levels for different velocity levels.

| Velocity | Avg Abs Error | Median Abs Err |
|---|---|---|
| 0.01 | 0.082 | 0.063 |
| 0.05 | 0.091 | 0.062 |
| 0.10 - Base | 0.083 | 0.050 |
| 0.15 | 0.080 | 0.045 |
| 0.25 | 0.096 | 0.053 |
| 0.50 | 0.093 | 0.042 |

Table 24. Comparison of average higher and lower error level runs for different velocity levels.

| Velocity | Avg of > .1 Error | Avg of < .1 Err | % < .1 Err |
|---|---|---|---|
| 0.01 | 0.164 | 0.041 | 0.669 |
| 0.05 | 0.198 | 0.043 | 0.693 |
| 0.10 - Base | 0.192 | 0.036 | 0.702 |
| 0.15 | 0.207 | 0.035 | 0.737 |
| 0.25 | 0.223 | 0.030 | 0.659 |
| 0.50 | 0.222 | 0.022 | 0.645 |

Increasing velocity correlates with lower average error indicating that a network can be successful trained at faster speeds and also suggesting that networks could, in some circumstances, potentially benefit from additional training beyond the 100 epochs tested. Notably, the increased training velocity causes a higher average error level in the high-error networks while reducing the average error level of the lower-error networks, as shown in Table 24. This explains the different trends shown in the average and median error values in Table 23. From the forgoing, it appears that increasing training velocity causes the network to move more quickly towards extremes resulting in increased and reduced average error levels in the higher-error and lower-error networks, respectively.

*6.5. Comparison of the Performance of Different Rule-Fact Configurations*

This section considers the impact of the number of rules and facts on network performance. Two different questions are answered. First, the impact on error and processing time of different sizes is considered. Also, the impact of having the same number of facts and rules versus more facts than rules is evaluated. Tables 25 to 27 present data for networks ranging in size from 50 facts and 50 rules to 250 facts and 250 rules.

Table 25. Comparison of average results and processing time for networks with different rule-fact configurations.

| Facts | Rules | Truth Network | Algorithm Network |
|---|---|---|---|
| 50 | 50 | 0.510 | 0.521 |
| 100 | 50 | 0.511 | 0.517 |
| (B) 100 | 100 | 0.522 | 0.521 |
| 150 | 100 | 0.514 | 0.534 |
| 150 | 150 | 0.489 | 0.503 |
| 200 | 150 | 0.486 | 0.496 |
| 200 | 200 | 0.512 | 0.515 |
| 250 | 200 | 0.516 | 0.528 |
| 250 | 250 | 0.488 | 0.498 |

Table 26. Comparison of average error levels for networks with different rule-fact configurations.

| Facts | Rules | Avg Abs Error | Median Abs Err |
|---|---|---|---|
| 50 | 50 | 0.083 | 0.054 |
| 100 | 50 | 0.068 | 0.032 |
| (B) 100 | 100 | 0.083 | 0.050 |
| 150 | 100 | 0.088 | 0.050 |
| 150 | 150 | 0.091 | 0.062 |
| 200 | 150 | 0.085 | 0.064 |
| 200 | 200 | 0.089 | 0.061 |
| 250 | 200 | 0.089 | 0.060 |
| 250 | 250 | 0.083 | 0.054 |

Table 27. Comparison of average higher and lower error level runs for networks with different rule-fact configurations.

| Facts | Rules | Avg of > .1 Error | Avg of < .1 Err | % < .1 Err |
|---|---|---|---|---|
| 50 | 50 | 0.198 | 0.038 | 0.713 |
| 100 | 50 | 0.168 | 0.027 | 0.709 |
| (B) 100 | 100 | 0.192 | 0.036 | 0.702 |
| 150 | 100 | 0.191 | 0.032 | 0.651 |
| 150 | 150 | 0.201 | 0.041 | 0.686 |
| 200 | 150 | 0.187 | 0.042 | 0.704 |
| 200 | 200 | 0.192 | 0.037 | 0.668 |
| 250 | 200 | 0.187 | 0.037 | 0.652 |
| 250 | 250 | 0.185 | 0.038 | 0.690 |

When comparing the data, there does not appear to be a clear pattern of error based on network size or symmetry versus asymmetry. While the training, processing and evaluation time increases with the increase in the size of the network, the asymmetric networks require less time than the symmetric networks with either the same number of facts or the same number of rules. Across the different run sets, the asymmetric runs typically take approximately an order of magnitude less time to operate than the symmetric ones.

Thus, it is clear that the proposed technique works with networks of a variety of different sizes. While 100 epochs of training performed sufficiently, it is expected that additional training may be beneficial for larger network sizes. Most notably, it is shown that there is a significant performance speed benefit to the use of networks with asymmetric numbers of rules and facts.

7. Impact of Network Learning Approaches

Up until this point, a single method of training, where a single target fact is used and a single starting fact has its value set initially (in addition to setting random values for all facts in the network) has been used. However, other approaches to training can also be used. These are discussed in this section.

Specifically, in addition to this approach of training a single path with the same facts, a path can be trained by using random facts which are assigned for each training run, while maintaining the rule fact contribution weightings between training epochs and updating them through successive epochs of training. This approach allows the network-in-training to have its rule fact contribution weightings trained under multiple conditions and using more nuanced data.

Additionally, training can occur using multiple paths through the network by selecting different target facts and setting the initial value of different initial facts. Both approaches – using the same fact values and regenerating random values for each training – can be used with this multiple path approach.

All three of these approaches improve on the base (single path with same facts) approach that has been used throughout this paper, up until this point. Data characterizing their performance is presented in Tables 28 to 30.

Table 28. Comparison of average results and processing time for different network learning approaches.

|  | Avg Result | Avg Result |
|---|---|---|
| Path - Same Facts (Base) | 0.522 | 0.521 |
| Path - Random Facts | 0.469 | 0.463 |
| Multiple Paths - Same Facts | 0.425 | 0.418 |
| Multiple Paths - Random Facts | 0.436 | 0.445 |

Table 29. Comparison of average error levels for different network learning approaches.

|  | Avg Abs Error | Median Abs Err |
|---|---|---|
| Path - Same Facts (Base) | 0.083 | 0.050 |
| Path - Random Facts | 0.078 | 0.044 |
| Multiple Paths - Same Facts | 0.054 | 0.013 |
| Multiple Paths - Random Facts | 0.052 | 0.022 |

Table 30. Comparison of average higher and lower error level runs for different network learning approaches.

|                               | Avg of > .1 Error | Avg of < .1 Err | % < .1 Err |
|---|---|---|---|
| Path - Same Facts (Base)      | 0.192 | 0.036 | 0.702 |
| Path - Random Facts           | 0.201 | 0.028 | 0.561 |
| Multiple Paths - Same Facts   | 0.190 | 0.020 | 0.795 |
| Multiple Paths - Random Facts | 0.180 | 0.022 | 0.814 |

The three alternate training techniques show performance benefits, over the base technique, in terms of error levels. The path-based random fact training has slightly less average error as compared to the base approach, and the multiple paths approaches both have approximately 60% of the average error. The impact on median error is more pronounced, with the median error for the multiple paths approaches being below half of the base approach's and the path-based random facts approach showing a 10% error reduction. The multiple paths techniques also have more lower-error level runs than the base approach. Given the foregoing, these approaches clearly outperform the base approach in terms of accuracy.

While these additional approaches provide an effective theoretical solution for training-based optimization of the networks, they have practical implementation implications. The path-based random facts approach presumes that fact values would vary significantly (random generation assumed that facts were equally likely to be any value within the range). This is an assumption that would likely not be true for many application areas, where fact values would be similar even with variation from instance to instance.

The multiple path-based approaches present a logistical issue, in that numerous data collection mechanisms would be required to collect data related to the various paths throughout the network. In a real-world environment, this may be highly impractical and may increase the data collection expense of training significantly.

Thus, while these approaches present theoretical benefit in network training, their practical utility may be limited in many application areas. Techniques that collect data from a limited number of paths or use multiple fact value sets, as opposed to a single path or single fact value set, may provide a level of error reduction in between the approaches in this section. Application of the proposed technique to application areas and its evaluation in these areas as well as the exploration of hybrid training techniques remain key areas for future work.

9. Conclusions and Future Work

This paper has presented a machine learning-based artificial intelligence technique based on the enhancement of expert system rule-fact networks. Because the system is created using only logically-valid and meaning-defined connections, the potential for the system learning a completely invalid correlation due to confounding relationships or data accuracy issues is minimized.

The use of the underlying rule-fact network requires a process to create it. Certainly, human creation is possible, as has been demonstrated with the creation of numerous expert systems described in Section 2. However, this process is time consuming and may introduce error. The system is highly reliant on the integrity of this rule-fact network and the experimentation presented herein has shown that the network itself provides a significant amount of the system's overall accuracy. While some robustness to network error and unnecessary network augmentation has been shown in simulation, the impact of error in real world applications (as well as its impact on the accuracy of rule-level weightings and intervening fact values) remains a key area for future study.

The impact of a variety of system operation choices such as training velocity, training epochs and network size has been evaluated. The system was shown to operate under a myriad of different conditions, with some configurations producing enhanced performance while others had little impact or impaired performance. The characterization of the impact of these operation choices may be informative with regards to the potential performance of the system for specific application areas.

Finally, three additional modes of training were demonstrated that exhibited enhanced performance. Two of the three, which used different paths through the rule-fact network, would also have the benefit of training the network more robustly for numerous potential uses (which don't relate to a single trained path). However, while these approaches showed enhanced performance, they present potential logistical issues and, in the case of approaches using fact value randomization, may rely on an assumption (of fact value variation) that is not valid in many application areas.

The work presented herein, thus, demonstrates the potential efficacy of the use of gradient descent-style training on an expert system and discusses the benefits that it may have for some applications. Future work will be needed to explore these benefits and to demonstrate the efficacy of the system in actual application areas. Additionally, methods to develop rule-fact networks that reduce the level of human effort required and techniques to prevent training from getting stuck in local minima are both key areas of needed future work.

References

[1] D.K. Malhotra, K. Malhotra, R. Malhotra, Evaluating Consumer Loans Using Machine Learning Techniques, in: Emerald Publishing Limited, 2020: pp. 59–69. https://doi.org/10.1108/s0276-897620200000020004.
[2] A. Deeks, The Judicial Demand for Explainable Artificial Intelligence, Columbia Law Rev. 119 (2019) 1829–1850.
[3] N. Stobbs, D. Hunter, M. Bagaric, Can sentencing be enhanced by the use of artificial intelligence?, Crim. Law J. 41 (2017) 261–277. https://eprints.qut.edu.au/115410/ (accessed February 24, 2021).
[4] V. Chiao, Predicting Proportionality: The Case for Algorithmic Sentencing, Crim. Justice Ethics. 37 (2018) 238–261. https://doi.org/10.1080/0731129X.2018.1552359.
[5] T. Gillespie, Content moderation, AI, and the question of scale, Big Data Soc. 7 (2020) 205395172094323. https://doi.org/10.1177/2053951720943234.
[6] G. Coppersmith, R. Leary, P. Crutchley, A. Fine, Natural Language Processing of Social Media as Screening for Suicide Risk., Biomed. Inform. Insights. 10 (2018) 1178222618792860. https://doi.org/10.1177/1178222618792860.

[7] X. Zhou, Y. Li, W. Liang, CNN-RNN Based Intelligent Recommendation for Online Medical Pre-Diagnosis Support, IEEE/ACM Trans. Comput. Biol. Bioinforma. (2020) 1–1. https://doi.org/10.1109/tcbb.2020.2994780.

[8] R. Caruana, A. Niculescu-Mizil, An empirical comparison of supervised learning algorithms, in: ACM Int. Conf. Proceeding Ser., ACM Press, New York, New York, USA, 2006: pp. 161–168. https://doi.org/10.1145/1143844.1143865.

[9] X. Goldberg, Introduction to semi-supervised learning, Synth. Lect. Artif. Intell. Mach. Learn. 6 (2009) 1–116. https://doi.org/10.2200/S00196ED1V01Y200906AIM006.

[10] H.B. Barlow, Unsupervised Learning, Neural Comput. 1 (1989) 295–311. https://doi.org/10.1162/neco.1989.1.3.295.

[11] T. Araujo, N. Helberger, S. Kruikemeier, C.H. de Vreese, In AI we trust? Perceptions about automated decision-making by artificial intelligence, AI Soc. 35 (2020) 611–623. https://doi.org/10.1007/s00146-019-00931-w.

[12] C. O'Neil, Weapons of Math Destruction, Broadway Books, New York, NY, USA, 2016.

[13] C. D'Ignazio, L.F. Klein, Data Feminism, MIT Press, Cambridge, MA, 2020.

[14] S.U. Noble, Algorithms of Oppression: How Search Engines Reinforce Racism Paperback, NYU Press, New York, NY, USA, 2018.

[15] G.S. Cole, Tort Liability for Artificial Intelligence and Expert Systems, Comput. Law. J. 10 (1990). https://heinonline.org/HOL/Page?handle=hein.journals/jmjcila10&id=131&div=&collection= (accessed October 19, 2020).

[16] J.K.C. Kingston, Artificial Intelligence and Legal Liability, in: Int. Conf. Innov. Tech. Appl. Artif. Intell., Springer International Publishing, 2016: pp. 269–279. https://doi.org/10.1007/978-3-319-47175-4_20.

[17] D. Waterman, A guide to expert systems, Addison-Wesley Pub. Co., Reading, MA, 1986.

[18] V. Zwass, Expert system, Britannica. (2016). https://www.britannica.com/technology/expert-system (accessed February 24, 2021).

[19] R.K. Lindsay, B.G. Buchanan, E.A. Feigenbaum, J. Lederberg, DENDRAL: A case study of the first expert system for scientific hypothesis formation, Artif. Intell. 61 (1993) 209–261. https://doi.org/10.1016/0004-3702(93)90068-M.

[20] O. Arsene, I. Dumitrache, I. Mihu, Expert system for medicine diagnosis using software agents, Expert Syst. Appl. 42 (2015) 1825–1834.

[21] B. Abu-Nasser, Medical Expert Systems Survey, Int. J. Eng. Inf. Syst. 1 (2017) 218–224. https://papers.ssrn.com/sol3/papers.cfm?abstract_id=3082734 (accessed January 17, 2021).

[22] E. Styvaktakis, M.H.J. Bollen, I.Y.H. Gu, Expert system for classification and analysis of power system events, IEEE Trans. Power Deliv. 17 (2002) 423–428.

[23] M. Pantic, L.J.M. Rothkrantz, Expert system for automatic analysis of facial expressions, Image Vis. Comput. 18 (2000) 881–905.

[24] J.M. McKinion, H.E. Lemmon, Expert systems for agriculture, Comput. Electron. Agric. 1 (1985) 31–40. https://doi.org/10.1016/0168-1699(85)90004-3.

[25] M. Kuehn, J. Estad, J. Straub, T. Stokke, S. Kerlin, An expert system for the prediction of student performance in an initial computer science course, in: IEEE Int. Conf. Electro Inf. Technol., 2017. https://doi.org/10.1109/EIT.2017.8053321.

[26] S. Kalogirou, Expert systems and GIS: An application of land suitability evaluation, Comput. Environ. Urban Syst. 26 (2002) 89–112. https://doi.org/10.1016/S0198-

9715(01)00031-X.
[27] J.M. Renders, J.M. Themlin, Optimization of Fuzzy Expert Systems Using Genetic Algorithms and Neural Networks, IEEE Trans. Fuzzy Syst. 3 (1995) 300–312. https://doi.org/10.1109/91.413235.
[28] S. Sahin, M.R. Tolun, R. Hassanpour, Hybrid expert systems: A survey of current approaches and applications, Expert Syst. Appl. 39 (2012) 4609–4617. https://doi.org/10.1016/j.eswa.2011.08.130.
[29] L.A. Zadeh, Fuzzy sets, Inf. Control. 8 (1965) 338–353. https://doi.org/10.1016/S0019-9958(65)90241-X.
[30] S. Mitra, S.K. Pal, Neuro-fuzzy expert systems: Relevance, features and methodologies, IETE J. Res. 42 (1996) 335–347. https://doi.org/10.1080/03772063.1996.11415939.
[31] W.A. Sandham, D.J. Hamilton, A. Japp, K. Patterson, Neural network and neuro-fuzzy systems for improving diabetes therapy, in: Institute of Electrical and Electronics Engineers (IEEE), 2002: pp. 1438–1441. https://doi.org/10.1109/iembs.1998.747154.
[32] A. Chohra, A. Farah, M. Belloucif, Neuro-fuzzy expert system E_S_CO_V for the obstacle avoidance behavior of intelligent autonomous vehicles, Adv. Robot. 12 (1997) 629–649. https://doi.org/10.1163/156855399X00045.
[33] E.P. Ephzibah, V. Sundarapandian, A Neuro Fuzzy Expert System for Heart Disease Diagnosis, Comput. Sci. Eng. An Int. J. 2 (2012) 17–23. https://citeseerx.ist.psu.edu/viewdoc/download?doi=10.1.1.1052.1581&rep=rep1&type=pdf (accessed February 22, 2021).
[34] S. Das, P.K. Ghosh, S. Kar, Hypertension diagnosis: A comparative study using fuzzy expert system and neuro fuzzy system, in: IEEE Int. Conf. Fuzzy Syst., 2013. https://doi.org/10.1109/FUZZ-IEEE.2013.6622434.
[35] B.A. Akinnuwesi, F.M.E. Uzoka, A.O. Osamiluyi, Neuro-Fuzzy Expert System for evaluating the performance of Distributed Software System Architecture, Expert Syst. Appl. 40 (2013) 3313–3327. https://doi.org/10.1016/j.eswa.2012.12.039.
[36] R. Rojas, The Backpropagation Algorithm, in: Neural Networks, Springer Berlin Heidelberg, Berlin, 1996: pp. 149–182. https://doi.org/10.1007/978-3-642-61068-4_7.
[37] R. Battiti, Accelerated Backpropagation Learning: Two Optimization Methods, Complex Syst. 3 (1989) 331–342. https://www.complex-systems.com/abstracts/v03_i04_a02/ (accessed February 22, 2021).
[38] B. Kosko, K. Audhkhasi, O. Osoba, Noise can speed backpropagation learning and deep bidirectional pretraining, Neural Networks. 129 (2020) 359–384. https://doi.org/10.1016/j.neunet.2020.04.004.
[39] W. Liu, L. Chen, Y. Chen, W. Zhang, Accelerating Federated Learning via Momentum Gradient Descent, IEEE Trans. Parallel Distrib. Syst. 31 (2020) 1754–1766. https://doi.org/10.1109/TPDS.2020.2975189.
[40] H.A. Abbass, Speeding Up Backpropagation Using Multiobjective Evolutionary Algorithms, Neural Comput. 15 (2003) 2705–2726. https://doi.org/10.1162/089976603322385126.
[41] C. Aicher, N.J. Foti, E.B. Fox, Adaptively Truncating Backpropagation Through Time to Control Gradient Bias, in: Proc. 35th Uncertain. Artif. Intell. Conf., MLR Press, 2020: pp. 799–808. http://proceedings.mlr.press/v115/aicher20a.html (accessed February 22, 2021).
[42] L. Chizat, F. Bach, Implicit Bias of Gradient Descent for Wide Two-layer Neural Networks Trained with the Logistic Loss, Proc. Mach. Learn. Res. 125 (2020) 1–34.

http://proceedings.mlr.press/v125/chizat20a.html (accessed February 22, 2021).

[43] J.F. Kolen, J.B. Pollack, Backpropagation is Sensitive to Initial Conditions, Complex Syst. 4 (1990) 269–280.

[44] P. Zhao, P.Y. Chen, S. Wang, X. Lin, Towards query-efficient black-box adversary with zeroth-order natural gradient descent, ArXiv. 34 (2020) 6909–6916. https://doi.org/10.1609/aaai.v34i04.6173.

[45] Z. Wu, Q. Ling, T. Chen, G.B. Giannakis, Federated Variance-Reduced Stochastic Gradient Descent with Robustness to Byzantine Attacks, IEEE Trans. Signal Process. 68 (2020) 4583–4596. https://doi.org/10.1109/TSP.2020.3012952.

[46] J.N.D. Gupta, R.S. Sexton, Comparing backpropagation with a genetic algorithm for neural network training, Omega. 27 (1999) 679–684. https://doi.org/10.1016/S0305-0483(99)00027-4.

[47] S. Basterrech, S. Mohammed, G. Rubino, M. Soliman, Levenberg - Marquardt training algorithms for random neural networks, Comput. J. 54 (2011) 125–135. https://doi.org/10.1093/comjnl/bxp101.

[48] A. Saffaran, M. Azadi Moghaddam, F. Kolahan, Optimization of backpropagation neural network-based models in EDM process using particle swarm optimization and simulated annealing algorithms, J. Brazilian Soc. Mech. Sci. Eng. 42 (2020) 73. https://doi.org/10.1007/s40430-019-2149-1.

[49] S. Park, T. Suh, Speculative Backpropagation for CNN Parallel Training, IEEE Access. 8 (2020) 215365–215374. https://doi.org/10.1109/ACCESS.2020.3040849.

[50] C. Lee, S.S. Sarwar, P. Panda, G. Srinivasan, K. Roy, Enabling Spike-Based Backpropagation for Training Deep Neural Network Architectures, Front. Neurosci. 14 (2020) 119. https://doi.org/10.3389/fnins.2020.00119.

[51] M. Mirsadeghi, M. Shalchian, S.R. Kheradpisheh, T. Masquelier, STiDi-BP: Spike time displacement based error backpropagation in multilayer spiking neural networks, Neurocomputing. 427 (2021) 131–140. https://doi.org/10.1016/j.neucom.2020.11.052.

[52] O. Beaumont, J. Herrmann, G. Pallez, A. Shilova, Optimal memory-aware backpropagation of deep join networks, Philos. Trans. R. Soc. A Math. Phys. Eng. Sci. 378 (2020). https://doi.org/10.1098/rsta.2019.0049.

[53] S. Kim, B.C. Ko, Building deep random ferns without backpropagation, IEEE Access. 8 (2020) 8533–8542. https://doi.org/10.1109/ACCESS.2020.2964842.

[54] W.D. Kurt Ma, J.P. Lewis, W.B. Kleijn, The HSIC bottleneck: Deep learning without back-propagation, in: Proc. AAAI Conf. Artif. Intell., arXiv, 2020: pp. 5085–5092. https://doi.org/10.1609/aaai.v34i04.5950.

[55] M. Gori, A. Tesi, On the problem of local minima in backpropagation, IEEE Trans. Pattern Anal. Mach. Intell. 14 (1992) 76–86. https://doi.org/10.1109/34.107014.

[56] M. Bianchini, M. Gori, M. Maggini, On the Problem of Local Minima in Recurrent Neural Networks, IEEE Trans. Neural Networks. 5 (1994) 167–177. https://doi.org/10.1109/72.279182.